\documentclass[10pt,twocolumn,letterpaper]{article}

\usepackage[pagenumbers]{cvpr} 

\usepackage{graphicx}
\usepackage{amsmath}
\usepackage{amssymb}
\usepackage{booktabs}
\usepackage{multirow}

\usepackage[pagebackref,breaklinks,colorlinks]{hyperref}

\usepackage[capitalize]{cleveref}
\crefname{section}{Sec.}{Secs.}
\Crefname{section}{Section}{Sections}
\Crefname{table}{Table}{Tables}
\crefname{table}{Tab.}{Tabs.}

\def\cvprPaperID{*****} 
\def\confName{CVPR}
\def\confYear{2022}

\begin{document}

\title{Facial Affective Behavior Analysis Method for 5th ABAW Competition}

\author{Shangfei Wang, Yanan Chang, Yi Wu, Xiangyu Miao, Jiaqiang Wu, Zhouan Zhu, Jiahe Wang, Yufei Xiao
\thanks{All the authors have equal contributions for the competition.}\\
University of Science and Technology of China\\
}

\maketitle

\begin{abstract}
Facial affective behavior analysis is important for human-computer interaction. 5th ABAW competition includes three challenges from Aff-Wild2 database. Three common facial affective analysis tasks are involved, i.e. valence-arousal estimation, expression classification, action unit recognition. For the three challenges, we construct three different models to solve the corresponding problems to improve the results, such as data unbalance and data noise. For the experiments of three challenges, we train the models on the provided training data and validate the models on the validation data.
\end{abstract}

\section{Introduction}
\label{sec:intro}
Facial affective behaviors are important facial signals to convey the emotion and intention for others. It is significant to perform facial affective behavior analysis automatically. Like previous competitions~\cite{kollias2019deep,kollias2019expression,kollias2019face,kollias2020analysing,kollias2021affect,kollias2021analysing,kollias2021distribution,kollias2022abaw,kollias2022abawsyn}, the 5th ABAW competition~\cite{Kollias2023ABAWVE} is based in-the-wild database for constructing real facial affective behavior models, i.e. Affwild2 databases.

The competition includes three different challenges for facial affective behavior analysis based on Affwild2. Specifically, there are valence-arousal~(VA) estimation, facial expression recognition, and facial action unit~(AU) recognition tasks, respectively. We design and train different models for three tasks independently. The data imbalance problem and data noise are common for the three tasks. We have to solve the diverse problems for different tasks respectively.

For the first challenge, VA estimation, we design a fusion method to combine three different backbone networks. The predictions are based on the contacted features. The relations between valence and arousal are also explored.

For the second challenge, expression classification, we utilize the ensemble learning method to improve the robustness of the trained model. The final predictions are based on the votes of multiple sub-classifiers. 

For the third challenge, AU recognition, we follow the previous work~\cite{nguyen2022ensemble} to leverage deep CNN and Transformer network to exploit the temporal and spatial information jointly.

The next sections will introduce the detailed methods and corresponding results for three challenges.


\section{Method and Results for VA Estimation}
\label{sec:va}
\begin{figure*}
	\centering
	\includegraphics[width=.8\textwidth]{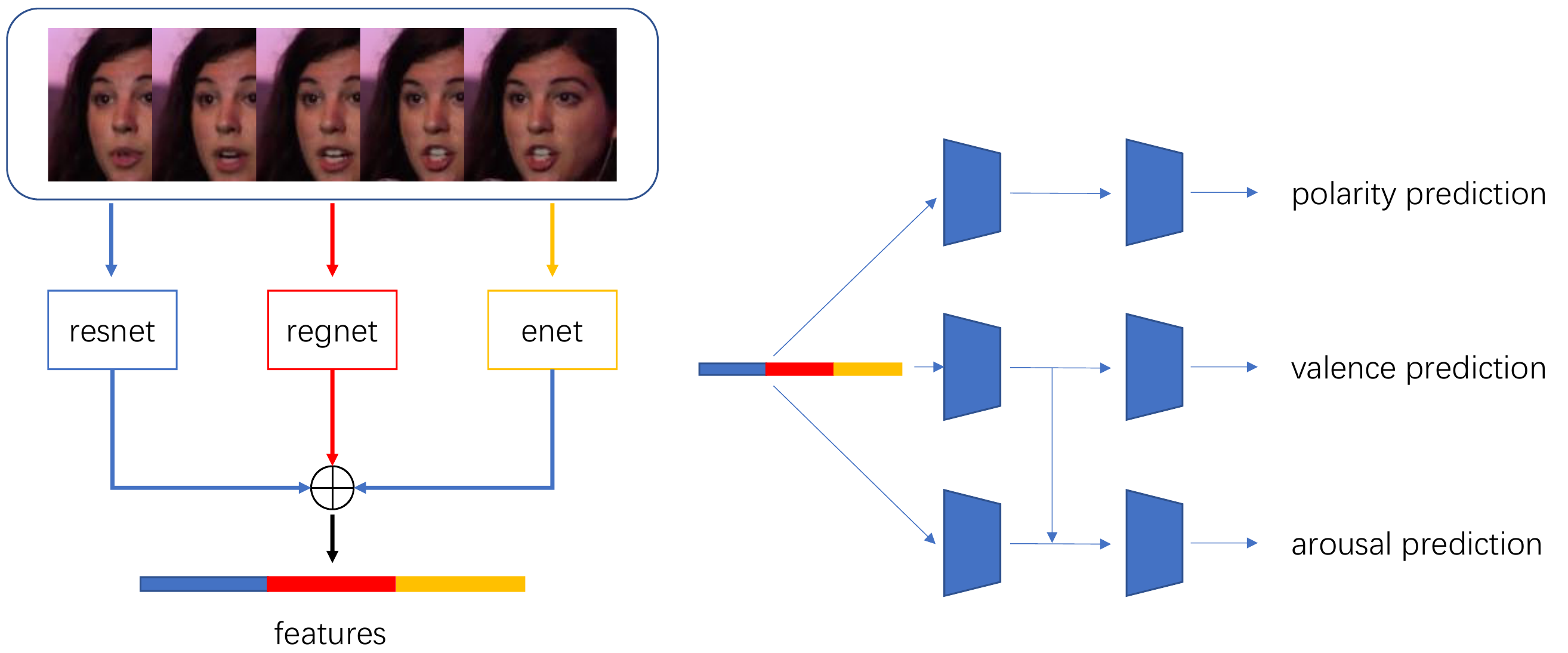} 
	\caption{The Framework of VA Prediction} 
	\label{fig:va framework} 
\end{figure*}

In the VA Estimation Challenge, given an image sequence consisting of $\{F^1, F^2, \dots, F^n\}$ from video X, the goal is to predict the label $y^v$ for valence and the label $y^a$ for arousal.

\subsection{Dataset Analysis}
In this challenge, after filtering out invalid images, whose labels are -5, 1680135 and 382021 images are used for training and testing, respectively. The value of $y^v$ and $y^a$ trend to positive. It is worth noting that part of the images has extreme labels like 1 or -1. Meanwhile, in the training set, it seems that as $|y^v|$ increases to 1, so does $y^a$. Arousal value is not low when valence value comes to a high level.
\subsection{Network}
The framework is shown in Figure \ref{fig:va framework}. Here Resnet50\cite{He_2016_CVPR}, Regnet\cite{radosavovic2020regnet}, and Efficientnet\cite{tan2019efficientnet} are used as the backbone of our framework. Images are input into these networks respectively and the extracted features are contacted together. Then, two polarity classifiers are used to judge if the labels are 1 or -1 for valence and arousal respectively. These extreme images will not be input into the next classifiers. Two classifiers are designed to predict valence and arousal value respectively. To further explore the connection between valence and arousal, the valence features are also input into the arousal classifier with the arousal features.
\subsection{Loss Function}
Now we introduce the loss function for each part.

To turn the regression task into a classification task, we first discrete the continuous labels. Apart from the images whose labels equal -1 or 1, the rest are considered as one class. The new label we call is $y^p$. Then the cross entropy is used. The loss is shown in eq \ref{eq:va ce_p}.
\begin{equation}
	CE_P = -\frac{1}{N}\sum_{i=1}^{N}(y^plog(x_i)),
	\label{eq:va ce_p}
\end{equation}
where $x_i$ is the output of the polarity classifiers and the loss is for valence and arousal respectively. The polarity classifiers are trained before training the valence and arousal classifiers.

The Concordance Correlation Coefficient (CCC) loss is used for valence and arousal prediction, which can be formulated as follows:
\begin{equation}
	CCC = 1-\frac{2\rho\sigma_{\hat{Y}} \sigma_{Y}}{\sigma_{\hat{Y}}^2+\sigma_{Y}^2 + (\mu_{\hat{Y}} - \mu_Y)^2},
	\label{eq:va ccc}
\end{equation}
where $\mu_Y$ is the mean of the labels and $\mu_{\hat{Y}}$ is the mean of the predictions. $\sigma_{Y}$ and $\sigma_{\hat{Y}}$ are the corresponding standard deviations, $\rho$ is the pearson correlation coefficient between labels and predictions.
\subsection{Experiment}
We adopt Adam optimizer to optimize the model and the learning rate is set to 0.0001. The results on the validation set are shown in the table \ref{table:va result}.
\begin{table}[h!]
	\begin{center}
		\caption{The CCC of Valence And Arousal In Validation Set.}
		\label{table:va result}
		\begin{tabular}{ccc} 
			\hline
			method & valence & arousal \\
			\hline
			baseline & 22.0 & 24.0\\
			ours & 25.7 & 38.3\\
			\hline
		\end{tabular}
	\end{center}
	
\end{table}

\section{Method and Results for Expression Classification}
\label{sec:exp}

\begin{figure*}[htbp]
	\centering
	\includegraphics[width=.8\textwidth]{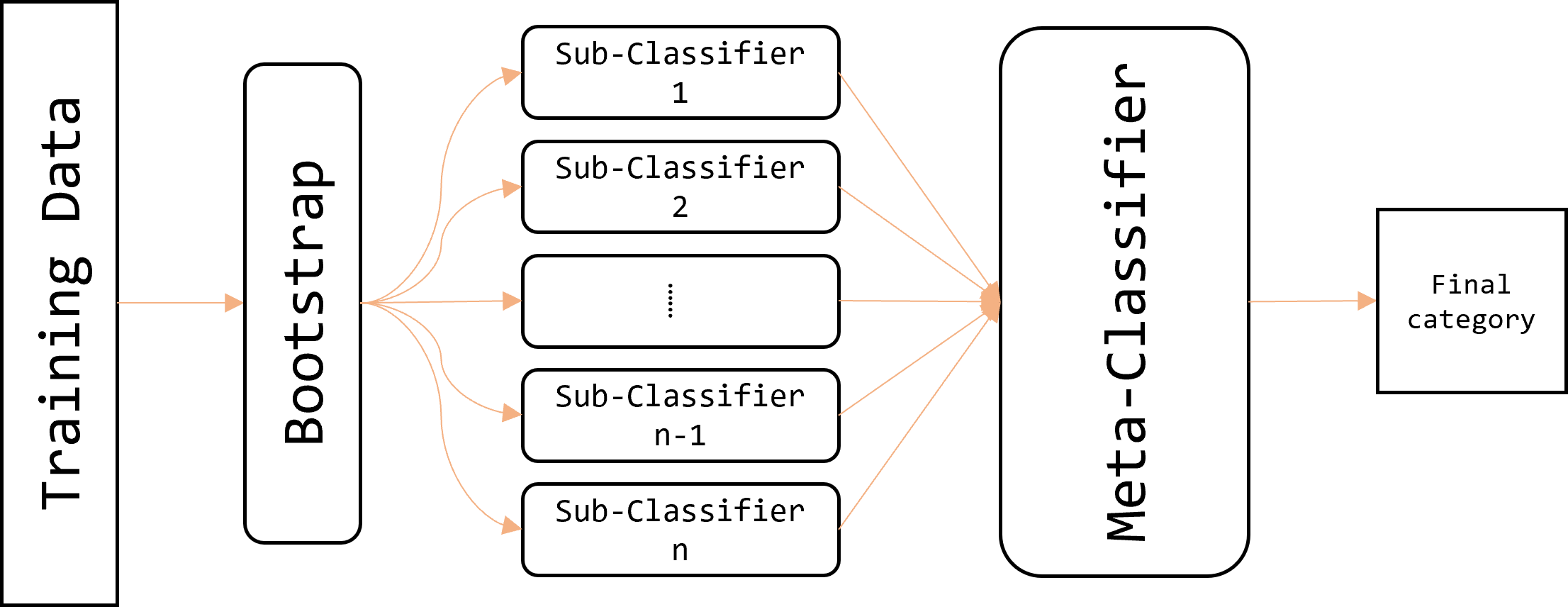}
	\caption{Method framework}
	\label{fig:frame}
\end{figure*}
\subsection{Overview}
In order to achieve accurate recognition of facial expressions under the requirements of this task, we need to learn a robust expression recognition model by using only expression labels. Therefore, we use the ensemble learning method. Specifically, we use the so-called Bagging method, which extracts a certain number of samples from the original training samples using the sampling method with put back, and train different sub-classifiers with different samples extracted. Finally, all the sub-classifiers predict the input image, and then pass the output to the Meta-classifier. The Meta-classifier votes according to certain rules to get the final category.

\subsection{Data Processing}
We delete all data labeled -1 from the dataset. For the sample labeled 7, we have two processing schemes: 1) The sample labeled 7 is regarded as a separate category and directly participates in the training; 2) Delete the samples labeled 7 from the training set, and only use the samples labeled 0-6 for training, and set a threshold during the prediction. When the maximum prediction probability of a certain sample is lower than the threshold, the sample will be classified into class 7. In addition, data enhancement operations such as Random Rotation, Random Resized Crop, Random Horizontal Flip and Color Jitter were applied to all training data.

\subsection{Model Architecture}
As shown in Figure \ref{fig:frame}, the architecture of our method consists of two stages: the first stage is the Sub-Classifier stage, and the second stage is the Meta-Classifier stage. 1) In the first stage, n groups of samples were extracted from all training data to train n Sub-Classifiers respectively. 2) In the second stage, the prediction results of all Sub-Classifiers are sent to the Meta-Classifier, and the final category is given by the Meta-Classifier according to certain rules.

\subsection{Ensemble Learning}
\subsubsection{Sub-Classifier stage}
In this stage, we found that Bootstrap selected all the data each time to get the best results. Therefore, each Sub-Classifier receives all the training data for learning.

\subsubsection{Meta-Classifier stage}
In this stage, the Meta-Classifier receives the predicted results from all Sub-Classifiers and gives the final category according to the following rules. 1) If the prediction label given by at least one Sub-Classifier is 1,2,3, or 5, the final category is the category given by the Sub-Classifier. 2) The order of priority among the four categories is: $2 > 3 > 5 > 1$. 3) If there is no Sub-Classifier with prediction label 1, 2, 3, 5, the final category is obtained by majority vote. It should be noted that Meta-Classifier is only used in prediction, not in training.

\subsubsection{Loss Function}
We used the following loss function in training the Sub-Classifier:
\begin{align*}
    L_{C}=L_{CE}+\lambda L_{Dice}
\end{align*}
Where $L_{CE}$ is the cross-entropy Loss function, $\lambda$ is the linear combination factor, and $L_{Dice}$ is the Dice Loss, and the formula is as follows:
\begin{align*}
    L_{Dice}=1-\frac1N \sum_{i=1}^N\frac{2TP_i}{2TP_i+FP_i+FN_i}
\end{align*}
Here $TP_i$ represents the number of True positives for class $i$ samples, $FP_i$ represents the number of False positives for class $i$ samples, and $FN_i$ represents the number of False negatives for class $i$ samples.

\subsection{Experiments}
\subsubsection{Dataset}
The Aff-Wild2 database\cite{kollias2019deep,kollias2019face,kollias2021affect,kollias2021distribution,kollias2022abaw,kollias2019expression,kollias2020analysing,kollias2021analysing,zafeiriou2017aff} is audiovisual (A/V) and in total consists of 548 videos of around 2.7M frames that are annotated in terms of the 6 basic expressions (i.e., anger, disgust, fear, happiness, sadness, surprise), plus the neutral state, plus a category 'other' that denotes expressions/affective states other than the 6 basic ones. Specifically, the database includes 177,498 neutral, 16,573 anger, 10,810 disgust, 9,080 fear, 95,633 happiness, 79,862 sadness,31,637 surprise and 165,866 other expressions.

\subsubsection{Training Details}
All of our experiments are on the Pytorch framework. Adam optimization is used to update the weights. The learning rate is $0.0005$. Our models are trained with the epoch of 20 and save the best performance on the validation set. We used the resnet18\cite{He_2016_CVPR}, resnet152\cite{He_2016_CVPR}, mobilenet\_v2\cite{sandler2018mobilenetv2}, densenet121\cite{huang2017densely} ,densenet201\cite{huang2017densely} model and loaded the pre-training parameter as our Sub-Classifier.  In this challenge, the final result is evaluated across the average F1 score of 8 emotion categories:
\begin{align*}
    F_1^{\text{final}} = \frac{\sum F_1^{\text{exp}}}{8}
\end{align*}
where $ F_1^{\text{exp}}$ is $F_1$ score of each expression.

\subsubsection{Results}
Table~\ref{table:exp} shows the results of expression classification on the validation set.

\begin{table}[!ht]
	\centering
	\begin{tabular}{ccc}
		\hline
		\textbf{Classifier} & \textbf{Acc} & \textbf{f1} \\ \hline
		Sub-Classifier 1 & 0.4454 & 0.2890 \\ 
		Sub-Classifier 2 & 0.4462 & 0.2833 \\ 
		Sub-Classifier 3 & 0.4321 & 0.2792 \\ 
		Sub-Classifier 4 & 0.4332 & 0.2941 \\ 
		Sub-Classifier 5 & 0.4367 & 0.2857 \\ 
		Meta-Classifier & 0.4618 & 0.3024 \\ \hline
	\end{tabular}
	\caption{Expression classification on the Affwild2 validation set. }
    \label{table:exp}
\end{table}

\section{Method and Results for AU recognition}
\label{sec:au}
In this section, we introduce our method for the Action Unit (AU) Detection Challenge in the 5th Workshop and Competition on Affective Behavior Analysis in-the-wild.

\subsection{Pre-processing}
Detecting AUs in videos first requires separating each frame of the video. Next, the face detection tool is used to detect facial landmarks, which typically include the eyebrows, eyes, nose, and mouth. The faces in each frame are then cropped according to the position of the face frame and normalized and aligned using the detected facial landmarks. We used the cropped and aligned images provided by the official ABAW competition.

We discovered that among the AU labels assigned to the data, those that were present were labeled as 1, and those that were absent were labeled as 0. However, some AUs were labeled as -1 when the faces were not well recognized. To prevent this data from affecting the training process, we excluded these samples from our analysis.

\subsection{Feature Extraction}
We employed various backbone networks to extract visual features. The first network used for facial feature extraction was the classical ResNet-50 \cite{He_2016_CVPR}. We tested the ResNet-50 with randomly initialized parameters and the pre-trained ResNet-50. The output dimension of the ResNet-50-based visual feature module was 2048.

We also employed RegNet \cite{radosavovic2020regnet} as a model for facial feature extraction. We used RegNet trained with the ImageNet dataset to extract visual features in both fixed network parameters and fine-tuned ways. We also attempted to randomly initialize RegNet and participate in subsequent training. However, due to GPU memory limitations, we could only use two versions of RegNet, RegNet-400MF and RegNet-800MF. The output dimension of the visual feature module based on RegNet-400MF was 440, and the output dimension of the visual feature module based on RegNet-800MF was 784.

\subsection{Architectures}
We followed the approach outlined in \cite{nguyen2022ensemble} and connected the temporal model to the backbone network. The feature sequence passes through three pipelines. In the first pipeline, the feature sequence is connected to a Transformer block to extract timing information and then to a fully connected layer with an output dimension of 12. In the second and third pipelines, the feature sequence is first up-sampled and then down-sampled before being connected to the fully connected layer or the Transformer block followed by the fully connected layer. The outputs of the three fully connected layers are then fused together for AU detection.

\subsection{Loss Function}
The evaluation metric for the AUs required for this competition is the F1 score. However, because there is an imbalance in the sample categories in the dataset, we decided to use focal loss to give more attention to the hard-to-classify samples and thereby improve the model's performance on the imbalanced data.
$$FL(p_t) = -\alpha_t (1 - p_t)^\gamma \log(p_t)$$
where $p_t$ represents the probability that the model predicts the sample belongs to the positive class. $\alpha_t$ denotes the weight assigned to the positive and negative samples, and $\gamma$ is a hyperparameter that adjusts the weights of the hard and easy samples.

\subsection{Experiments}
\subsubsection{Dataset}
There are 548 videos in Aff-Wild2, labeled with 12 AUs: AU1, AU2, AU4, AU6, AU7, AU10, AU12, AU15, AU23, AU24, AU25, and AU26. The ABAW3 competition provided 295 of these videos as the training set, 105 as the validation set, and 141 as the test set. Additionally, the ABAW3 competition provided 145,273 frames for the training set, 27,087 frames for the validation set, and 51,245 frames for the test set, along with cropped and aligned face images.

\subsubsection{Experimental Setting}
The models were trained on an Nvidia GeForce GTX 3090 GPU with 24GB memory per GPU. The models were trained using the SGD optimizer with a learning rate of 0.7. They were trained for 24 epochs with a batch size of 16 and a dropout rate of 0.3. The image size was $112 \times 112$, and the sequence length was 256.

\subsubsection{Results}
Table \ref{table:AU} shows the experimental results of our proposed method on the aff-wild2 dataset validation set. The average F1 score was used as an evaluation metric for the Action Unit Detection task. As seen from the table, our method is competitive and outperforms the baseline results.

\begin{table}[t]
	\centering
	\begin{tabular}{lll}
		\hline
		method                      & feature      & F1     \\ \hline
		baseline                    &              & 0.39   \\ \hline
		\multirow{3}{*}{our method} & Resnet-50    & 0.5591 \\
		& Regnet-400mf & 0.6813 \\
		& Regnet-800mf & 0.6983 \\ \hline
	\end{tabular}
	\caption{Action unit detection on the Affwild2 validation set. }
	\label{table:AU}
\end{table}

\section{Conclusion}
In the paper, we have introduced the methods for three challenges of 5th ABAW competitions. The experimental results on the validation dataset are also included for facial valence and arousal estimation, action unit recognition, and expression classification.

{\small
\bibliographystyle{ieee_fullname}
\bibliography{egbib}
}

\end{document}